\documentclass{article}
\usepackage{ijcai24}

\usepackage{times}
\usepackage{soul}
\usepackage{url}
\usepackage[hidelinks]{hyperref}
\usepackage[utf8]{inputenc}
\usepackage[small]{caption}
\usepackage{graphicx}
\usepackage{utfsym}
\usepackage{amsmath}
\usepackage{amssymb}
\usepackage{amsthm}
\usepackage{makecell}
\usepackage{colortbl}
\usepackage{multirow}
\usepackage{multicol}
\usepackage{booktabs}
\usepackage{xcolor}
\usepackage{color}
\usepackage{tcolorbox}
\usepackage{subfig}
\usepackage{algorithm}
\usepackage{algorithmic}
\usepackage{tabularx}
\usepackage{enumitem}
\usepackage{afterpage}
\usepackage{natbib}

\usepackage[switch]{lineno}

\newcommand\our{\textsc{Alympics}}
\title{\our{}: LLM Agents meet Game Theory \\ Exploring Strategic Decision-Making with AI Agents}

\author{
Shaoguang Mao\thanks{\ \ Equal contributions.}
\and
Yuzhe Cai$^*$\thanks{\ \ This work was done during internship at MSRA.}\and
Yan Xia\and
\\
Wenshan Wu \and
Xun Wang \and
Fengyi Wang$^\dag$ \and
Tao Ge\thanks{\ \ Corresponding Author.} \And
Furu Wei
\affiliations
Microsoft Research Asia
\emails
\{shaoguang.mao,  v-yuzhecai, yanxia\}@microsoft.com,
\\
\{wenshan.wu, xunwang, v-fengyiwang, tage, fuwei\}@microsoft.com
}
\begin{document}

\maketitle

\begin{abstract}
This paper introduces \textit{Alympics} (Olympics for Agents), a systematic simulation framework utilizing Large Language Model (LLM) agents for game theory research. %~\cite{Myerson1991GameT}.
\textit{Alympics} creates a versatile platform for studying complex game theory problems, bridging the gap between theoretical game theory and empirical investigations by providing a controlled environment for simulating human-like strategic interactions with LLM agents. In our pilot case study, the "Water Allocation Challenge," we explore \textit{Alympics} through a challenging strategic game focused on the multi-round auction on scarce survival resources. This study demonstrates the framework's ability to qualitatively and quantitatively analyze game determinants, strategies, and outcomes. Additionally, we conduct a comprehensive human assessment and an in-depth evaluation of LLM agents in strategic decision-making scenarios. Our findings not only expand the understanding of LLM agents' proficiency in emulating human strategic behavior but also highlight their potential in advancing game theory knowledge, thereby enriching our understanding of both game theory and empowering further research into strategic decision-making domains with LLM agents. Codes, prompts, and all related resources are available at \href{https://github.com/microsoft/Alympics}{\emph{Alympics}}.
\end{abstract}

\begin{figure}[t]
  \centering
  \includegraphics[width=\columnwidth]{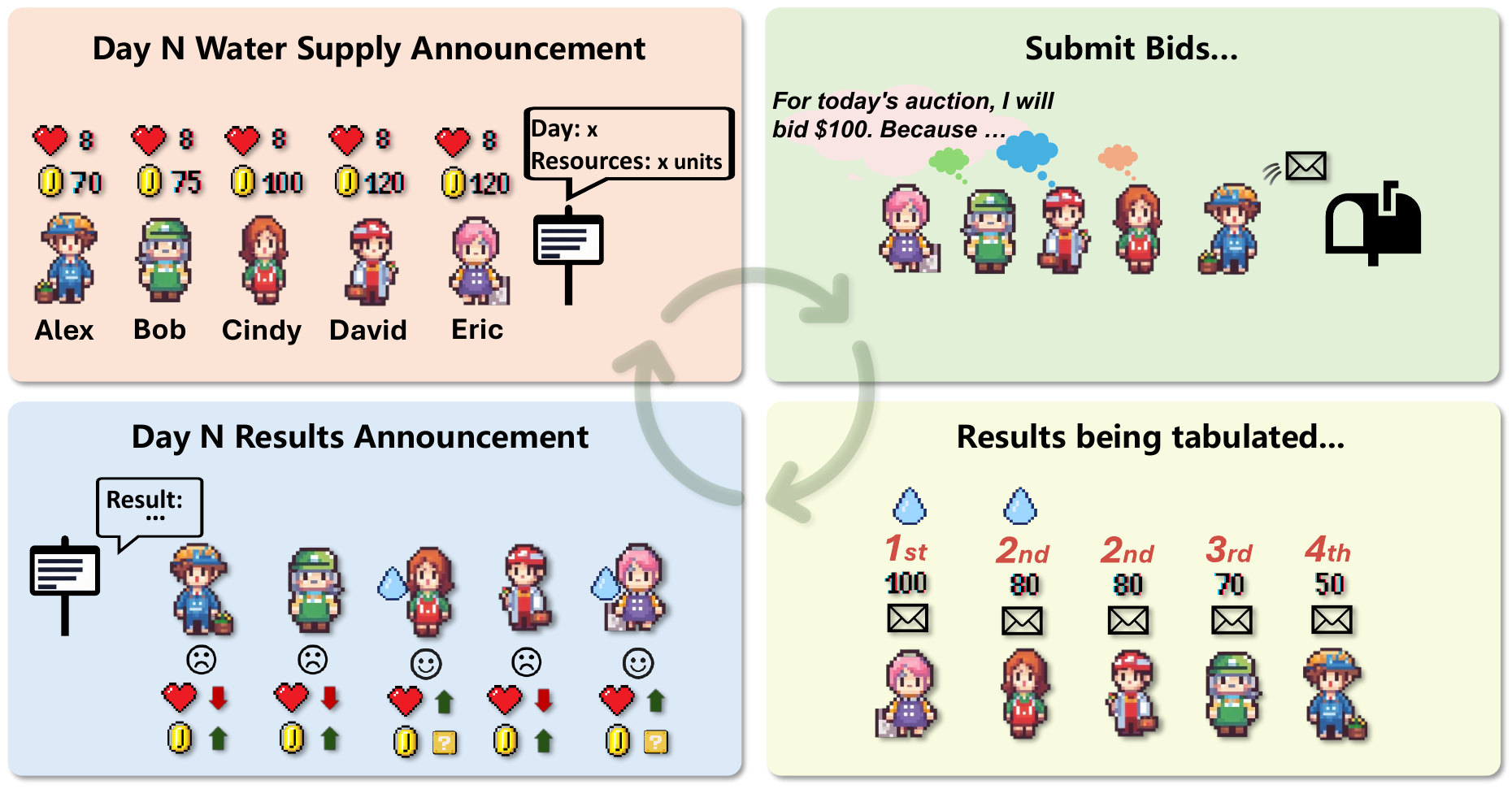}
  \caption{"Water Allocation Challenge". Players are tasked with ensuring survival over 20 days by strategically acquiring water resources through daily auctions. Each player has different income and different water demand. Daily water supply varies and will be announced before daily auction. To allocate water resources, a sealed-bid auction will be conducted daily. Acquiring water increases HP, while failing to do so decreases HP. Players whose HP drop below or equal to 0 will be eliminated from the game.}% This game incorporates elements of a series of classic games and also avoids potential data leakage issues that may occur in classic games.}
  \label{fig:game}
\end{figure}

\section{Introduction}
Game theory is a branch of mathematics that studies strategic interactions among rational agents. It has applications in many fields, such as economics \citep{shubik1981game, pohjola1986applications}, social sciences \citep{sanfey2007social, ziems2023can}, computer science \citep{yang2020overview}, and biology \citep{archetti2019cooperation}. However, the study of game theory in practice presents challenges: Many real-world problems in game theory cannot be solved through simple theoretical deductions. Instead, they often require real-world experiments, which can be expensive, time-consuming, and ethically complex due to the involvement of human participants.

%Many real-world game theory problems cannot be resolved through simple theoretical mathematical deductions; instead, they often require conducting real-world experiments, which can be costly, time-consuming, and ethically complex due to the involvement of human participants.

% 所幸如今有了新的转机...
Fortunately, recent advancements in Large Language Models (LLMs) \citep{openai2023gpt4, bubeck2023sparks, touvron2023llama} and LLM-based agents\citep{sumers2023cognitive, li2023camel, lin2023swiftsage, guo2023gpt} now offer a new opportunity to study these complex game theory problems with AI. These developments have enabled the creation of increasingly sophisticated systems capable of emulating human behavior in various dimensions, including style, tone, personality, emotions, and even collaborative and competitive efforts\citep{wang2023survey, talebirad2023multi, madaan2023self,wang2023does, de2023emergent, zhao2023competeai, park2023generative, chen2023put, abdelnabi2023llm, zhang2023exploring, lore2023strategic, horton2023large}. For example, \citet{xu2023exploring} illustrate this progress using the example of Werewolf, where they observe non-preprogrammed emergent strategic behaviors in LLMs during gameplay, such as trust, confrontation, camouflage, and leadership. However, there are still three open questions on using LLM and agent for game theory research: How to construct a unified, controllable, and efficient framework for simulating human strategic interactions and facilitating game theory research? What methods are available for conducting game theory research using the LLM Agent framework? Does the LLM Agent demonstrate strategic behavior akin to humans, and what level of LLM agent achieved in the strategic reasoning?

\begin{figure*}[t]
  \centering
  \includegraphics[width=0.98\textwidth]{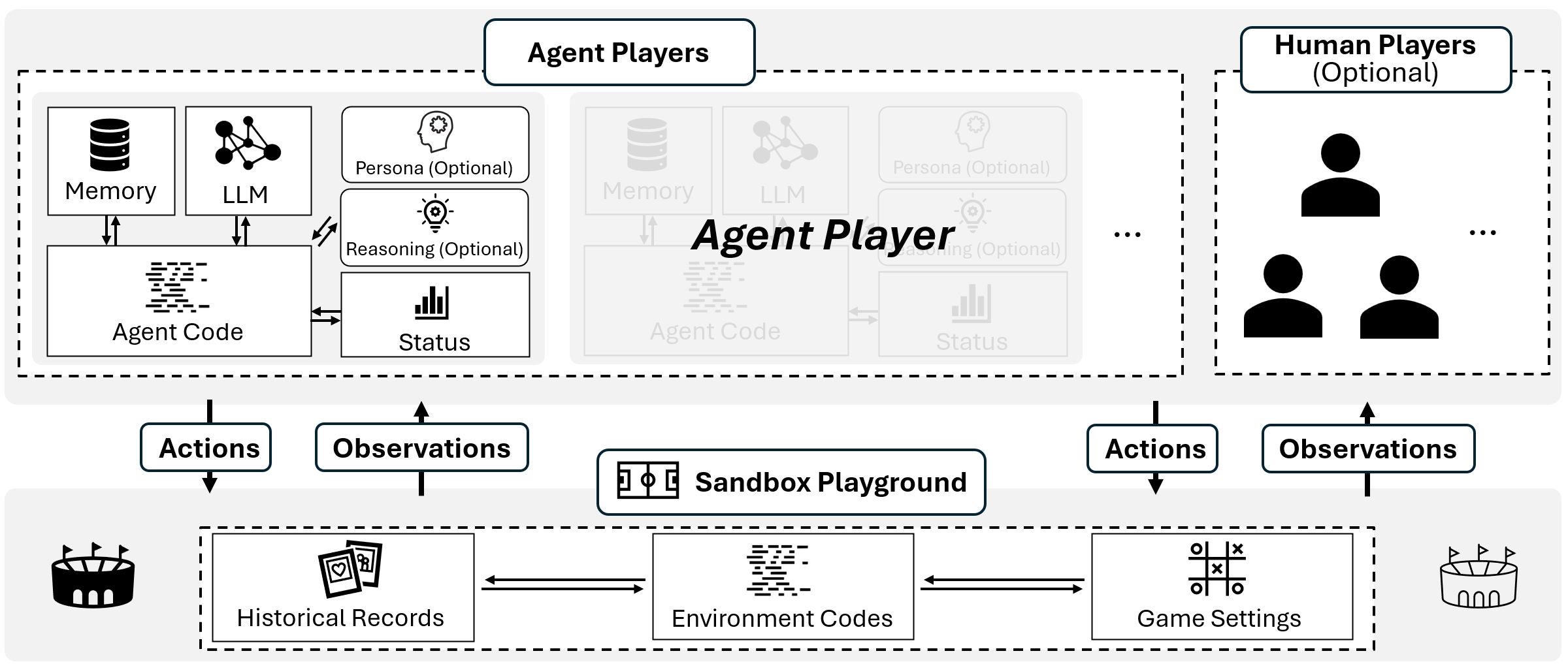}
  \caption{The architecture of \textit{Alympics} comprises the Sandbox Playground and Players.  The Sandbox Playground creates an environment where game settings, as specified by researchers, are executed. Agent players, along with the optional human players, actively engage in the game within this environment.}
  \label{fig:framework}
\end{figure*}

In this paper, we argue that LLMs can be used to implement pseudo-agents which can participate in game-theoretic scenarios and provide insights into the dynamics and outcomes of strategic interactions. We introduce \textit{Alympics}, Olympics for Agents, a new simulation framework for game theory using LLM agents. \textit{Alympics} incorporates a Sandbox Playground, Agent Players, and the option for Human Players, enabling the construction of realistic and dynamic models of human interactions. By leveraging the capabilities of LLM agents, our framework provides researchers with a controlled, scalable, and reproducible platform for exploring various game scenarios and testing hypotheses in game theory.

To exemplify the practicality and effectiveness of simulating and researching strategic decision-making scenarios, we present a pilot case study centered around an unequal competition for limited resources. As shown in Fig.\ref{fig:game}, this game is a reduction of a series of classic game theory problems such as auctions, dynamic games, and unequal competition. It also avoids potential data leakage issues that may occur in classic games. Through the manipulation of resource availability and participating agent personalities, we demonstrate how \textit{Alympics} can be employed to investigate the determinants influencing strategic decision-making and game outcomes.

Although there are many works on simulating human behaviors through language agents, it is still unclear whether the agents' simulations demonstrate rational reasoning and strategic behaviors. So we conduct an exhaustive human assessment of the agent's performance in game-theoretic scenarios. This involved evaluating aspects like \textbf{information utilization}, \textbf{logical reasoning}, \textbf{strategic effectiveness}, \textbf{adaptability}, and \textbf{long-term planning}, to determine the current level of agents in simulating human dynamic strategic behaviors. The evaluation results found that humans' perception of the machine's performance in games is similar to their self-assessment results. The result is crucial for judging conducting game-theoretic experiments through \textit{Alympics} or other AI agent settings. Our findings underscore the potential of LLM agents in deepening our comprehension of game theory and decision-making processes within intricate socioeconomic contexts.

In summary, this paper has the following contributions: (1) the proposal of a systematic LLM agent-based framework to facilitate game theory research, 
(2) The development of a game setting inspired by a range of classic game theory problems, showcasing Alympics's strength in both qualitative and quantitative analysis of game determinants, strategies, and outcomes.  
% (2) the demonstration of how \textit{Alympics} can explore game theory concepts, highlighting its utility in analyzing strategies and game outcomes, and 
(3) The comprehensive subject evaluation of LLM agents' performance in strategic scenarios, which reveals the capability of LLMs in mimicking complex human strategic behaviors in socioeconomic contexts. These contributions not only enhance our understanding of game theory but also hold the promise to influence research in AI agents across various domains where strategic decision-making is crucial.

\section{Alympics: An LLM Agent-based Game Theory Playground}
\textit{Alympics} is a systematic framework leveraging LLM agents for exploring game theory research. This framework comprises: Playground, Agent Players and Human Players (optional). As illustrated in the Figure.\ref{fig:framework}, Agent Players and Human Players engage in game on the Sandbox Playground within the defined game settings.

\subsection{Sandbox Playground}
The Sandbox Playground serves as the environment for conducting games, providing a versatile and controlled space for agent players interactions. It includes three key components: 

\textbf{Environment codes} define the rules and mechanics governing the game, ensuring a consistent and reliable framework for experimentation. 

\textbf{Historical records} maintain a comprehensive archive of past game records, enabling detailed analysis and facilitating the assessment of agent strategies over time. 

\textbf{Game settings} allow for the precise customization of parameters, offering researchers the flexibility to explore a wide range of scenarios. 

These components form a flexible and robust platform upon which Agent Players and optional Human Players engage in strategic interactions.

\subsection{Agent Players}
Agent Players constitute an indispensable component of the \textit{Alympics} framework, embodying LLM-powered agent entities that participate in strategic interactions within the Sandbox Playground. Each Agent Player is defined by the following key elements: 

\textbf{Agent Codes} represent the underlying algorithmic logic that controls \textit{decision-making} and \textit{strategy formulation}; 

\textbf{Player Status} encapsulates the current state and information accessible to the agent; 

\textbf{Large Language Model} is a powerful engine that augments the agent's cognitive capabilities and enables natural language interactions; 

\textbf{Memory Cache} provides a repository for storing and retrieving relevant historical information \citep{shinn2023reflexion, hu2023chatdb}; 

\textbf{Reasoning Plugin} offers specialized logic or algorithms for complex decision-making processes \citep{wei2022chain, yao2023tree}; 

\textbf{Persona Setting} defines the agent's behavioral profile and strategic inclinations \citep{wang2023unleashing, xu2023expertprompting};

\textbf{Other Components} include additional elements tailored to specific research needs, such as tool utilization\citep{shen2023hugginggpt, liang2023taskmatrix, qin2023toolllm} and augmentation.

These components equip Agent Players with the requisite intelligence and adaptability to engage in strategic gameplay, contributing to the dynamic landscape of game theory research within the \textit{Alympics} framework.

\section{Pilot Demonstration: Water Allocation Challenge}
\textit{Alympics} provides a research platform for conducting experiments on complex strategic gaming problems. As a pilot demonstration, we implemented a game called the 'Water Allocation Challenge'. This game incorporates elements of auction theory, resource allocation, survival strategy, repeated games, Nash equilibrium, fairness, and risk management. It represents characteristics of a series of classic games and also avoids potential data leakage issues that may occur in classic games.

\begin{figure*}
  \centering
  \includegraphics[width=0.99\textwidth]{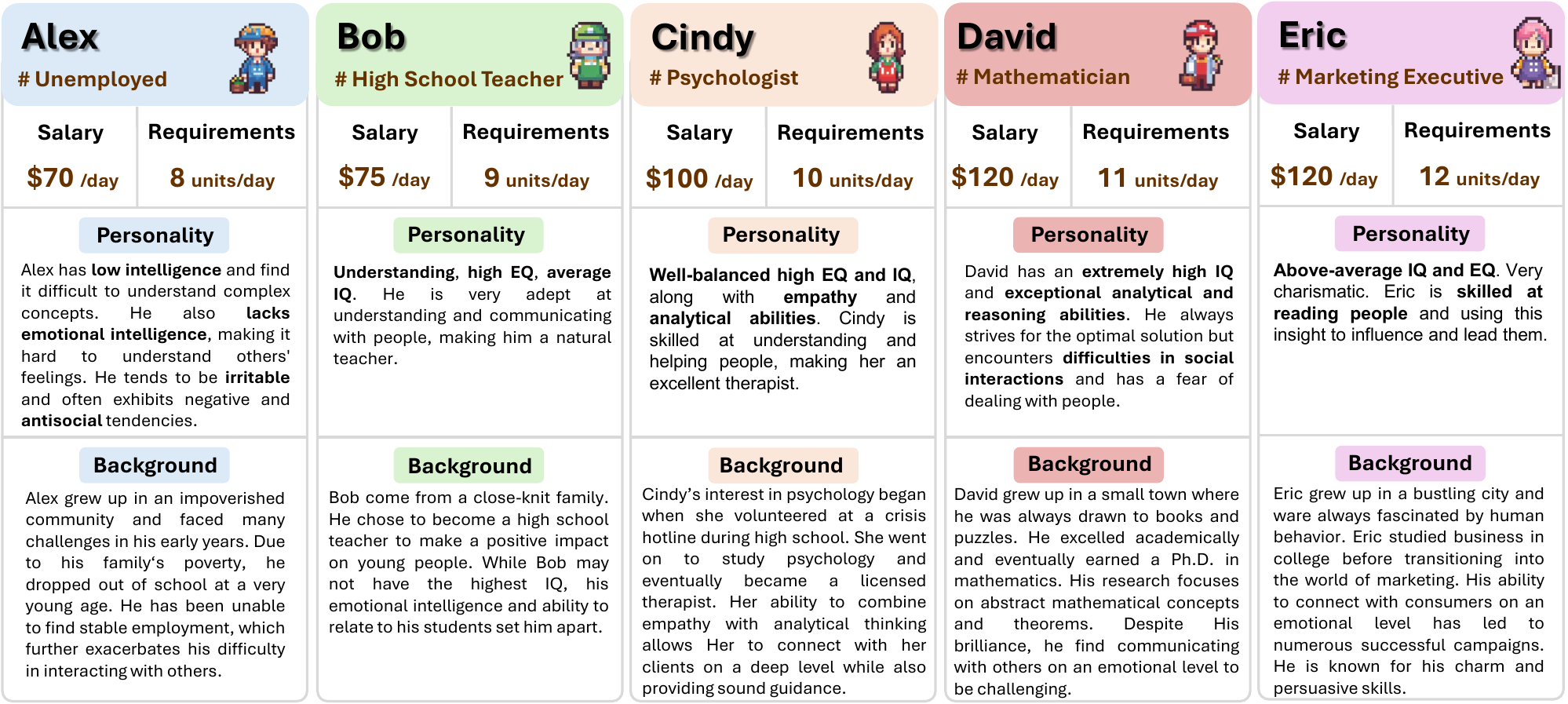}
  \caption{The player's information and persona. In all experiments, basic information (including name, daily salary and requirements) will be used. While Profession, Personality, and Background are only used in the Player Persona comparative experiments.}
  \label{fig:persona}
\end{figure*}

\subsection{Game Settings}
W Town is experiencing a rare drought. Every resident in W Town has been tasked with ensuring their survival over a period of 20 days by acquiring water resources. Each player will participate in daily auctions to bid for the necessary water resources to meet their individual needs. Here are the detailed game rules and settings:
\begin{itemize}
\item \textbf{Goal}: All residents share the same objective: to survive until the end of the 20-day period.

\item \textbf{Player Info}: Each player has unique water requirements and varying salaries. Refer to specific information in Figure \ref{fig:persona}.

\item \textbf{Health Points}: Each player has a maximum of 10 health points and starts with 8. If a player's health points drop to or below 0, they will be eliminated from the game.

\item \textbf{Routine}: Every day, all players will bid on water resources to meet their needs. If a player goes without obtaining water resources for a consecutive number of days (referred to as `No-Water Days') equal to `n', the player's health will be reduced by `n' points on that day. If their water needs are met, 2 points will be added to the player's health, and the count of No-Water Days will be reset to 0.

\item \textbf{Supply}: The daily water supply varies but is always less than the total demand. The specific amount will be announced before the daily auction.

\item \textbf{Auction Rule}: To allocate water resources, a sealed-bid auction will be conducted daily. Each resident submits a single bid for their entire water requirement. The government will allocate water resources based on the principle of the highest bidder until the remaining water resources are insufficient to meet anyone's requirement.

\item \textbf{Tie Rule}: In the event of a tie, priority will be given to residents with lower requirements.
\end{itemize}
\subsection{Game Analysis}
The Water Allocation Challenge presents an intriguing game theory scenario.

\textbf{Strategic Interactions} This game involves complex strategic interactions where players must consider not only their needs but also the behaviors and strategies of others.% This complexity provides fertile ground for studying various aspects of strategic decision-making.

\textbf{Learning and Adaptation} Players may adapt their strategies over time based on past experiences and observations of others' behavior.

\textbf{Uncertain Environments} The dynamic and uncertain nature of the game progress, like the varying daily supply, allows for the exploration of strategies under uncertain conditions.

\textbf{Inequality and Fairness} The inherent inequality among players, characterized by disparities in income and needs, presents an opportunity to study how players with different resources and requirements formulate strategies and interact.

The game has parallels with real-world scenarios involving resource allocation and competition. Conducting human experiments is costly, hard to control, and not easily reproducible. We leverage \textit{Alympics} for emulation in order to investigate the phenomenon of strategic interaction among agents. Insights gained from emulations could have implications for policy-making in areas where resources are scarce and need to be allocated efficiently.

\subsection{Research Topics and Methodology}
Our focus lies in the competitions among agent players within \textit{Alympics} and the qualitative assessment of these agents' strategic behaviors.

We repeat the emulations and analyze the mutual influence of bidding strategies among agents, the impacts of income inequality and variations in requirements on player survival, as well as the evolution of agent players' strategic adaptations. Then, we modify the emulation parameters, such as resource availability and agent characteristics, to study their influence on the agents' bidding behaviors.

To evaluate the extent to which LLM Agents demonstrate strategic reasoning and strategy evolution, we invited 10 human subjects to conduct a subjective evaluation of the agent's performance in the game. These findings offer insights into the agents' capabilities for strategic decision-making within complex socioeconomic environments. Details are elaborated in the Section \ref{sec:evaluation}.

\section{Experiments}
\subsection{Implementation}
GPT-4 is utilized for Sandbox Playground implementation. Meanwhile, Each agent player is equipped with an individual instance of GPT-4\footnote[1]{GPT-4-32k on Azure, Model version: 2023-07-01-preview}. 

Assume the system message as $S$ (i.e., game setting), bidding results as $B = [b_1, b_2, ..., b_{20}]$, where $b_n$ represents the bidding summary of round $n$. Additionally, consider bidding results from the $i$-th player as $R_i = [r^i_1, r^i_2, ..., r^i_{20}]$, where $r^i_n$ is the response from the $i$-th player in round $n$. Assume the participants' information denoted as $I = [i_1, i_2, ..., i_{20}]$, where $i_n$ represents the broadcasted information of all participants in round $n$, including health points, remaining budget, and consecutive No-Water Days. All prompts can be found in the appendix\ref{app:prompts}.

To obtain response $r^i_{n}$ from $i$-th player for a round $n$, the operation is as eq.\ref{eq:prompt}.
\begin{equation}
r^i_{n} = f(S, r^i_1, b_1, i_1, ..., r^i_{n-1}, b_{n-1}, i_{n-1})\label{eq:prompt}
\end{equation}
where $f$ stands for GPT-4.

\begin{table}[t]
\centering
\begin{tabular}{cccc}
\hline \textbf{Group}
& \textbf{ID} & \textbf{Resource Abundance} & \textbf{Persona} \\
\hline
& 1 & Low & \usym{2718} \\ 
(a) & 2 & Medium & \usym{2718} \\
& 3 & High & \usym{2718} \\
\hline
& 4 & Low & \usym{2714} \\
(b) & 5 & Medium & \usym{2714} \\
& 6 & High & \usym{2714} \\
\hline
\end{tabular}
\caption{Experimental Settings. In group (a), no persona is assigned to agent players, while in group (b), personas are assigned to agent players. In each group, there are three settings corresponding to low, medium, and high resource abundance respectively.}
\label{table:settings}
\end{table}

\subsection{Variables}
\textbf{Resource Abundance} We varied resource abundance in three conditions: Low, Medium, and High. Considering the total water demand from all agent players is 50 units, in the Low condition, the daily water supply follows a discrete uniform distribution ranging from 10 to 20. In the Medium condition, it follows a discrete uniform distribution ranging from 15 to 25. In the High condition, it follows a discrete uniform distribution ranging from 20 to 30.

We introduce the Resource Satisfaction Rate (${\rm RSR}$), representing the mathematical expectation of the resource's satisfaction rate for the total demand of surviving players. 
\begin{equation}
{\rm RSR}= \frac{\mathbb{E}({\rm resources})}{ {\textstyle \sum_{p\in {\rm survivors}}^{}{\rm requirment}_p} }\label{eq:RSR}
\end{equation}
The closer ${\rm RSR}$ is to 0, the more intense the current competition is. When ${\rm RSR}$ is greater than or equal to 1, it means that all players' demands can be fully satisfied, and it can be considered to there is no competition between players.

In low, medium, and high resource abundance settings, the ${\rm RSR}$ values are 0.3, 0.4, and 0.5 respectively.

\textbf{Player Persona} We compare versions without assigning persona settings to agent players (i.e., directly using GPT-4 to participate in the game) and versions where personas were assigned to agent players. Each persona setting contains three parts: \textbf{profession}, \textbf{personality}, and \textbf{background}. The agent players are assigned with distinct personas, including various professions, intelligence levels, and emotional intelligence levels in human society. By introducing personas, the heterogeneity among the agent players is further enhanced. Through comparative experiments, we aim to investigate whether assigning personas will affect the player's survival and strategy. The persona settings can be found in the Figure \ref{fig:persona}. 

\subsection{Experimental Settings}

We designed six experimental settings, as outlined in Table \ref{table:settings}. In Experimental Group (a), comprising settings 1 to 3, no persona is assigned to the agents. They are provided with low, medium, and high abundance resources, respectively. Experimental Group (b) includes settings 4 to 6, where each agent is assigned a persona (see Fig.\ref{fig:persona}). Similar to Group (a), agents in Group (b) are provided with low, medium, and high abundance resources. By comparing experiments within each group, we can observe the impact of resource abundance on player strategies and survival. Comparing Groups (a) and (b) allows us to observe the influence of persona assignment on player strategies and survival conditions.

For each setting, we conducted the experiment 10 times to obtain stable results. An example of one round record is shown in Appendix.\ref{app:example}.

\definecolor{myRed}{HTML}{FFB6C1}
\definecolor{myGreen}{HTML}{BCE672}
% \definecolor{myGreen}{HTML}{90EE90}
\definecolor{myDarkGreen}{HTML}{2E8B57}
\newcommand{\N}{\cellcolor{myRed}\textcolor{red}{\usym{2718}}}
\newcommand{\Y}{\cellcolor{myGreen}\textcolor{myDarkGreen}{\usym{2714}}}

\begin{table*}[th]
\centering
\resizebox{\textwidth}{!}{
\begin{tabular}{clcccccccccccccccccccccc}
\toprule
\textbf{R.A.} & \textbf{Player} & \multicolumn{11}{c}{\textbf{w/o Persona}} & \multicolumn{11}{c}{\textbf{w/ Persona}} \\
\midrule
\multirow{9}{*}{\textbf{Low}} & & \multicolumn{11}{c}{Setting 1} & \multicolumn{11}{c}{Setting 4} \\
\cmidrule(lr{1em}){3-13} \cmidrule(lr{1em}){14-24}
& & 1 & 2 & 3 & 4 & 5 & 6 & 7 & 8 & 9 & 10 & Avg. & 1 & 2 & 3 & 4 & 5 & 6 & 7 & 8 & 9 & 10 & Avg. \\
% \cline{2-22}
& Alex & \N & \N & \N & \N & \N & \N & \N & \N & \N & \Y & 0.10
& \Y & \Y & \N & \N & \N & \N & \Y & \N & \N & \N & 0.30 \\
& Bob & \N & \N & \N & \N & \Y & \N & \N & \N & \N & \Y & 0.10
& \N & \N & \N & \N & \N & \N & \Y & \N & \N & \N & 0.10 \\
& Cindy & \Y & \Y & \N & \N & \N & \Y & \Y & \Y & \N & \N & 0.50
& \N & \Y & \N & \Y & \N & \N & \Y & \N & \Y & \N & 0.40 \\
& David & \N & \Y & \Y & \Y & \N & \Y & \Y & \Y & \N & \Y & 0.70
& \N & \N & \Y & \Y & \Y & \Y & \N & \Y & \N & \Y & 0.60 \\
& Eric & \Y & \N & \Y & \N & \Y & \N & \N & \N & \Y & \N & 0.40
& \Y & \Y & \Y & \N & \Y & \Y & \N & \Y & \Y & \N & 0.70 \\
& RSR\_S & 0.30 & 0.30 & 0.30 & 0.30 & 0.30 & 0.30 & 0.30 & 0.30 & 0.30 & 0.30 & 0.30 & 0.30 & 0.30 & 0.30 & 0.30 & 0.30 & 0.30 & 0.30 & 0.30 & 0.30 & 0.30 & 0.30 \\
& RSR\_E & 0.68 & 0.71 & 0.65 & 1.36 & 0.71 & 0.71 & 0.71 & 0.71 & 1.25 & 0.79 & 0.83
& 0.75 & 0.50 & 0.65 & 0.71 & 0.65 & 0.65 & 0.56 & 0.65 & 0.68 & 1.36 & 0.72 \\
\midrule
\multirow{9}{*}{\textbf{Medium}} & & \multicolumn{11}{c}{Setting 2} & \multicolumn{11}{c}{Setting 5} \\
\cmidrule(lr{1em}){3-13} \cmidrule(lr{1em}){14-24}
& & 1 & 2 & 3 & 4 & 5 & 6 & 7 & 8 & 9 & 10 & Avg. & 1 & 2 & 3 & 4 & 5 & 6 & 7 & 8 & 9 & 10 & Avg. \\
% \cline{2-22}
& Alex & \N & \Y & \Y & \Y & \Y & \Y & \Y & \N & \Y & \Y & 0.80
& \N & \Y & \Y & \Y & \Y & \N & \Y & \Y & \N & \N & 0.60 \\
& Bob & \N & \Y & \N & \Y & \N & \N & \Y & \N & \Y & \Y & 0.50
& \N & \Y & \Y & \Y & \Y & \Y & \N & \N & \N & \Y & 0.60 \\
& Cindy & \Y & \Y & \Y & \Y & \Y & \Y & \Y & \Y & \N & \N & 0.80
& \Y & \Y & \N & \Y & \Y & \N & \Y & \N & \N & \Y & 0.60 \\
& David & \Y & \Y & \Y & \Y & \Y & \N & \N & \Y & \Y & \Y & 0.80
& \Y & \Y & \N & \N & \N & \Y & \N & \Y & \Y & \N & 0.50 \\
& Eric & \Y & \Y & \Y & \Y & \N & \Y & \Y & \Y & \Y & \Y & 0.90
& \N & \Y & \Y & \Y & \Y & \Y & \Y & \Y & \Y & \Y & 0.90 \\
& RSR\_S & 0.40 & 0.40 & 0.40 & 0.40 & 0.40 & 0.40 & 0.40 & 0.40 & 0.40 & 0.40 & 0.40 & 0.40 & 0.40 & 0.40 & 0.40 & 0.40 & 0.40 & 0.40 & 0.40 & 0.40 & 0.40 & 0.40 \\
& RSR\_E & 0.61 & 0.40 & 0.49 & 0.40 & 0.49 & 0.48 & 0.67 & 0.61 & 0.50 & 0.50 & 0.51
& 0.95 & 0.40 & 0.69 & 0.51 & 0.51 & 0.63 & 0.67 & 0.65 & 0.87 & 0.65 & 0.65 \\
\midrule
\multirow{9}{*}{\textbf{High}} & & \multicolumn{11}{c}{Setting 3} & \multicolumn{11}{c}{Setting 6} \\
\cmidrule(lr{1em}){3-13} \cmidrule(lr{1em}){14-24}
& & 1 & 2 & 3 & 4 & 5 & 6 & 7 & 8 & 9 & 10 & Avg. & 1 & 2 & 3 & 4 & 5 & 6 & 7 & 8 & 9 & 10 & Avg. \\
% \cline{2-22}
& Alex & \Y & \Y & \Y & \Y & \Y & \Y & \Y & \Y & \Y & \Y & 1.00
& \Y & \Y & \Y & \Y & \Y & \Y & \Y & \Y & \Y & \Y & 1.00 \\
& Bob & \Y & \Y & \Y & \Y & \Y & \Y & \Y & \Y & \Y & \Y & 1.00
& \N & \Y & \Y & \N & \Y & \Y & \Y & \Y & \Y & \Y & 0.80 \\
& Cindy & \Y & \Y & \Y & \Y & \Y & \Y & \Y & \Y & \Y & \Y & 1.00
& \Y & \Y & \Y & \Y & \Y & \Y & \Y & \Y & \Y & \Y & 1.00 \\
& David & \Y & \Y & \Y & \Y & \Y & \Y & \Y & \Y & \Y & \Y & 1.00
& \Y & \Y & \Y & \Y & \Y & \Y & \Y & \Y & \Y & \Y & 1.00 \\
& Eric & \Y & \Y & \Y & \Y & \Y & \Y & \N & \Y & \Y & \Y & 0.90
& \Y & \Y & \Y & \Y & \Y & \Y & \Y & \Y & \Y & \Y & 1.00 \\
& RSR\_S & 0.50 & 0.50 & 0.50 & 0.50 & 0.50 & 0.50 & 0.50 & 0.50 & 0.50 & 0.50 & 0.50 & 0.50 & 0.50 & 0.50 & 0.50 & 0.50 & 0.50 & 0.50 & 0.50 & 0.50 & 0.50 & 0.50 \\
& RSR\_E & 0.50 & 0.50 & 0.50 & 0.50 & 0.50 & 0.50 & 0.66 & 0.53 & 0.54 & 0.55 & 0.53
& 0.61 & 0.50 & 0.50 & 0.61 & 0.50 & 0.50 & 0.50 & 0.50 & 0.50 & 0.50 & 0.52 \\
\bottomrule
\end{tabular}}
\caption{Survival Status Records: The table lists the survival status of each player at the end of the games for all settings. A '$\checkmark$' indicates the player's survival at the end of the game, while a '$\times$' indicates the player's eliminated during the game. Based on the survival status, the table reports the Survival Rate for each player under different settings. Additionally, we report the Resource Satisfaction Rate (${\rm RSR}$) at the beginning (${\rm RSR}_{\rm S}$) and end of the game (${\rm RSR}_{\rm E}$). R.A. stands for Resource Abundance.}
\label{suvivalrecords}
\end{table*}

\subsection{Indicators}
We observe the following indicators in the experiment. 

${\rm RSR}_{\rm S}$ denotes the Resource Satisfaction Rate at the beginning of each game, while ${\rm RSR}_{\rm E}$ represents the Resource Satisfaction Rate at the end of the game. By comparing ${\rm RSR}_{\rm S}$ to ${\rm RSR}_{\rm E}$, we can observe the change in per capita resource allocation before and after the game. Additionally, by examining ${\rm RSR}_{\rm E}$, we can evaluate the level of resource abundance after each game.

We track the number of survivors, denoted as ${\rm N}_{\rm survivor}$, in each game as well as the survival rates (${\rm SR}$) of different players. For instance, ${\rm SR}_{\rm A}$ represents the survival rate of player A over 10 rounds of games under a specific setting.

Furthermore, we record the minimum successful bid price $p$ in each round. Here, $p_n$ represents the minimum successful bid price in round $n$. The variations in $p_n$ provide insights into the bidding strategies and trends of players.

\begin{figure*}[th]
\centering
\subfloat[Setting 1]{
\includegraphics[width=0.33\linewidth]{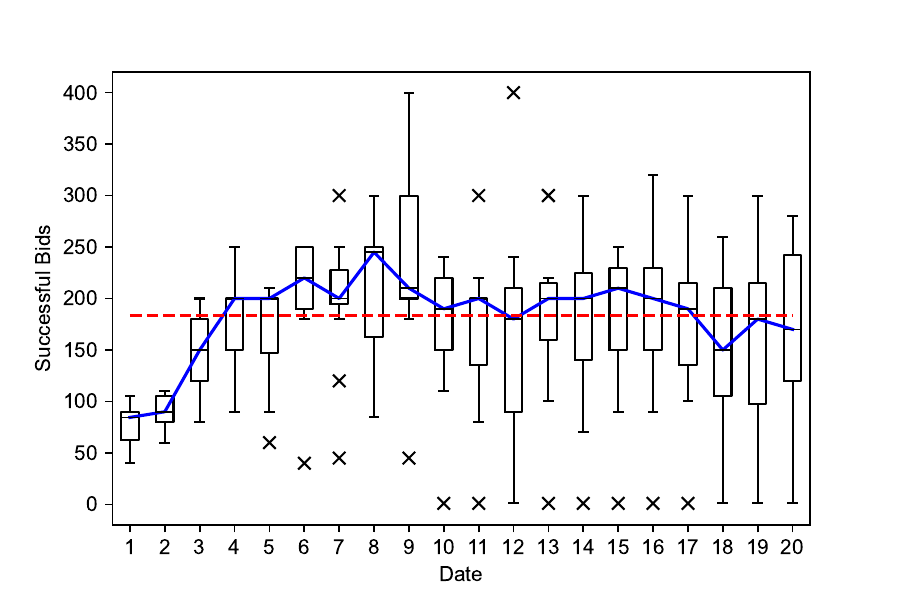}
}
\subfloat[Setting 2]{
\includegraphics[width=0.33\linewidth]{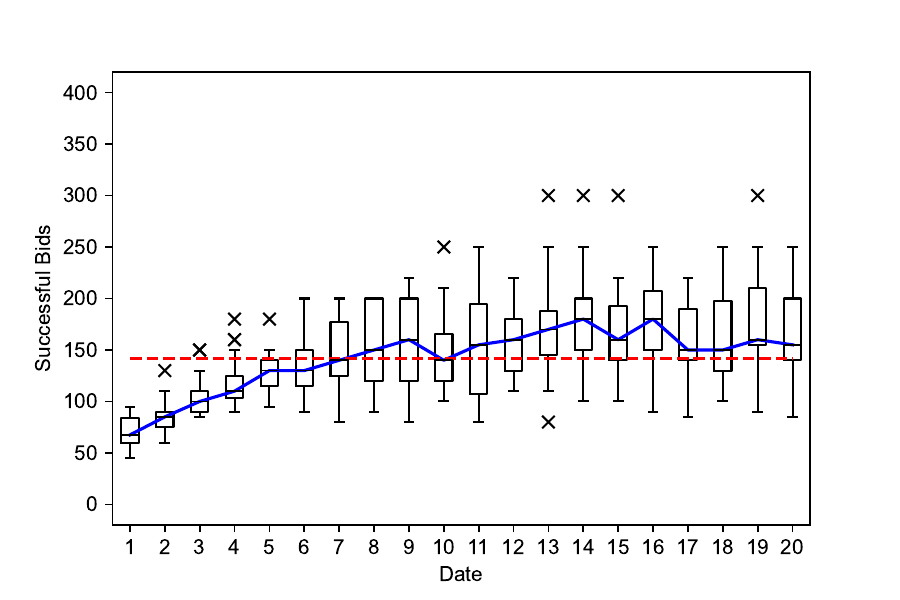}
}
\subfloat[Setting 3]{
\includegraphics[width=0.33\linewidth]{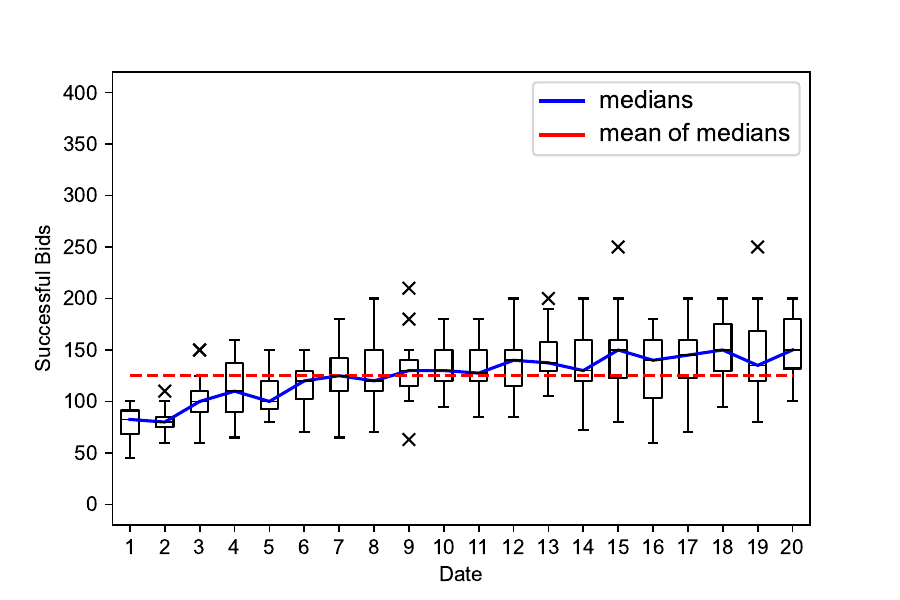}
}

\subfloat[Setting 4]{
\includegraphics[width=0.33\linewidth]{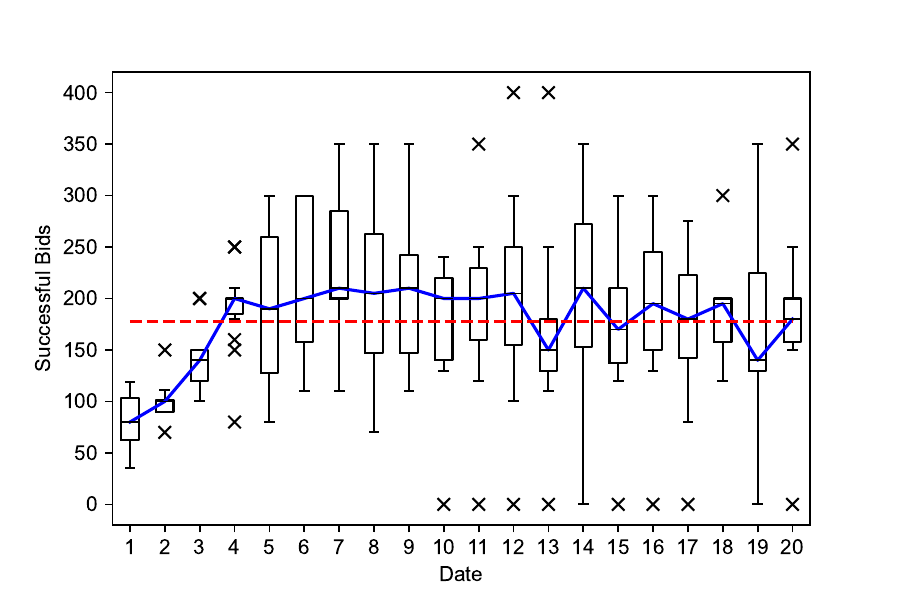}
}
\subfloat[Setting 5]{
\includegraphics[width=0.33\linewidth]{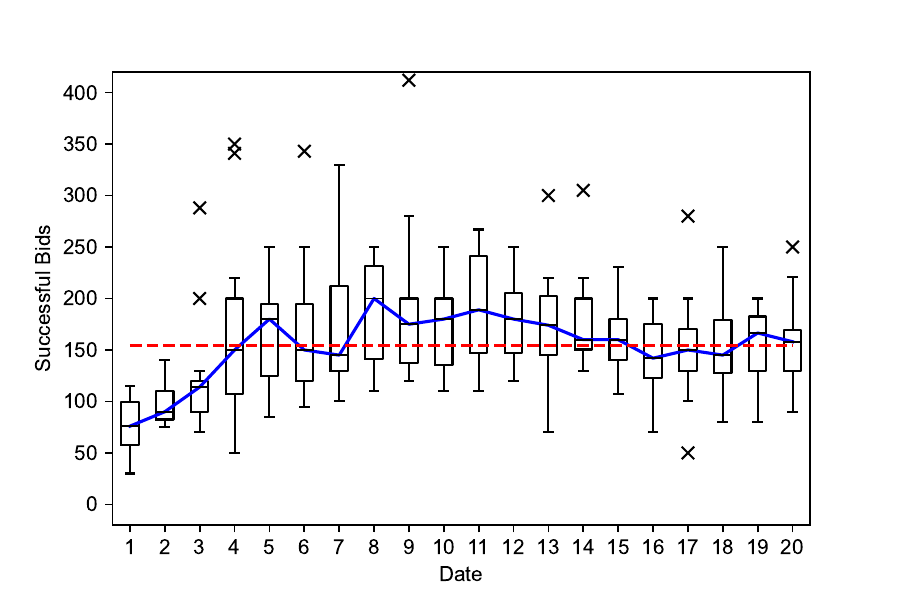}
}
\subfloat[Setting 6]{
\includegraphics[width=0.33\linewidth]{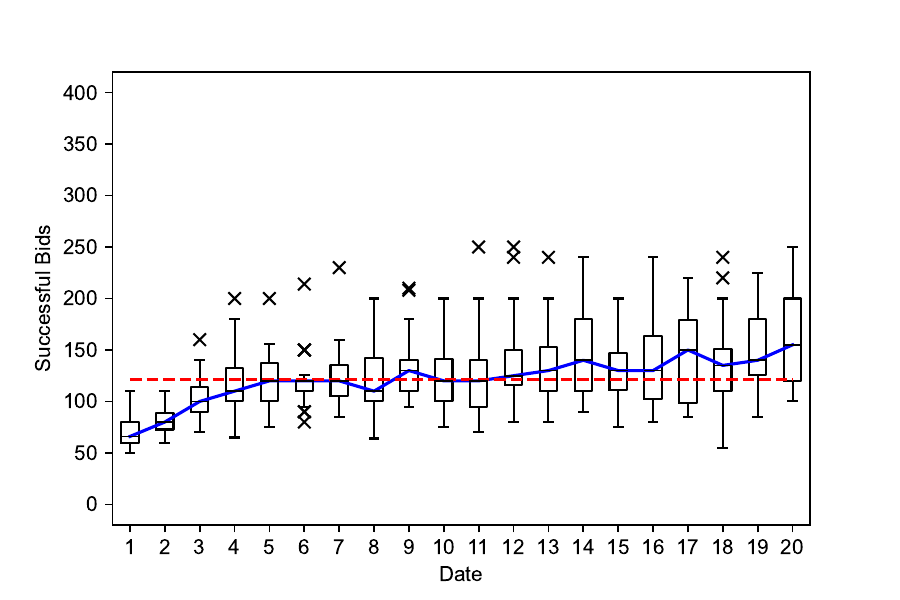}
}
\caption{Box Plots for bidding details of all settings. Subplots record the minimum successful bid for 10 independent experiments under corresponding setting. The x-axis represents the date, and the y-axis represents the price. These figures display the absolute value and trends in bids. Additionally, we have plotted the trend of the daily median with a blue line, and the average of the median for 20 game days with a red dashed line.}
\label{fig:bidboxplot}
\end{figure*}
% \begin{figure}[th]
% \centering
% \subfloat[Box Plot of ${\rm N}_{\rm survivor}$]{
%   \includegraphics[width=0.8\linewidth]{Figure/N.png}
% }

% \subfloat[Box Plot of ${\rm RSR}_{\rm E}$]{
% \includegraphics[width=0.8\linewidth]{Figure/RSR_E.png}
% }
% \caption{Box Plots of number of survivor ${\rm N}_{\rm survivor}$ and Resource Satisfaction Rate at the end-game ${\rm RSR}_{\rm E}$ in different experimental settings.}
% \label{fig:boxplot}
% \end{figure}

\section{Results}
\subsection{Survival Status}
Table.\ref{suvivalrecords} documents the survival status of each player in  experiments. The average ${\rm N}_{\rm survivor}$ in low resource abundance is significantly lower than medium and high resource abundance.

Also, there are significant differences in the survival rates among players. In section \ref{sec:analysis1}, we will provide a detailed analysis of the advantages and disadvantages of each player and their correlation with players' survival rates.

\subsection{Bidding Status}
Fig.\ref{fig:bidboxplot} records the bidding details of all settings. Specifically, each subplot is a box plot which shows the minimum successful bid $p$ in each round for 10 independent experiments under corresponding settings. 

Through the results, we draw the following conclusions:

\textbf{Absolute Bid}: As the abundance of resources increases, the minimum successful bid $p$ decreases. In conditions of abundant resources, survival is reasonably assured, leading players to commit less money to competition. Conversely, in conditions of resource scarcity, competition intensifies, prompting players to invest more money in survival.

\textbf{Bidding Trends}: For experiments with scarce resources (Experiments 1 and 4), the minimum successful bid initially rises rapidly, then decreases as the game progresses (after 10 days). However, for experiments with abundant resources (Experiments 3 and 6), there is a monotonic upward trend throughout the entire game, but with a more moderate increase. In games where survival is guaranteed, as the total accumulated wealth increases, inflation occurs.

\section{Analysis}
\subsection{Players' Advantages and Disadvantages}\label{sec:analysis1}
In the game setting, players differ in terms of their incomes and demands, which determine their advantages and disadvantages in the game.

The allocation rule dictates that the highest bidder wins, with priority given to players with lower demands in case of a tie. Therefore, in terms of monetary advantage, the hierarchy is Eric = David $>$  Cindy $>$  Bob $>$ Alex. While in terms of demand advantage, the order is Alex $>$ Bob $>$ Cindy $>$ David $>$ Eric. Depending on the varying availability of resources, player's advantages and disadvantages will be dynamically adjusted.

Through comparative experiments in setting 1-3, we observed significantly higher survival rates for Cindy, David, and Eric compared to Alex and Bob. Specifically, in experimental setting 1, the survival rate for Alex and Bob is only 0.10. This suggests that in settings without personalized characteristics, money plays a crucial role in survival. It's important to note that David and Eric have similar salaries, but David's daily demand is lower than Eric's. Considering tiebreaker conditions and the probability of resources meeting demand under low resource abundance, a player with low demand has an advantage over those with high demand.

This experiment involved repeatable experiments to draw significant conclusions. This underscores the importance of using our framework for game theory exploration, as it allows researchers to utilize LLM Agents for batch experiments, providing an empirical perspective to validate or challenge theories in game theory.

\subsection{Resource Abundance and Competition}
Intuitively, competition is more intense when resources are scarce. The results confirm this assumption. According to observations of  ${\rm N}_{\rm survivor}$, we can conclude that when the game starts with lower initial resource supply (${\rm RSR}$), the average player survival rate is lower.

Another interesting observation is that when the initial resource supply (${\rm RSR}_{\rm S}$) is lower, end-game resources (${\rm RSR}_{\rm E}$) are relatively more abundant. We notice that games that begin with intense competition lead players to adopt more aggressive strategies, whereas games starting with abundant resources lead players to adopt more conservative strategies.

\subsection{Persona and Survival}
Assigning personas increases the heterogeneity among agent players. Simultaneously, it enables agents to emulate the thinking patterns of various groups of people.

Compared to not assigning personas, the survival rate of players increases under conditions of low resource supply but decreases under conditions of medium resource supply. Additionally, we observed significant changes in survival rates for certain players before and after being assigned a persona. For instance, Cindy and David experienced a noticeable decrease in survival rates in the game, whereas player Eric's survival rate significantly improved. Investigating the reactions and survival conditions of players with different personas would be a very interesting direction.

\section{Subjective Evaluation}\label{sec:evaluation}
Although there are many works on simulating human behaviors through LLM agents, it is still unclear whether the agents' simulations demonstrate rational reasoning and strategic behaviors. This is an important question as it determines the usability of Agent's simulation in mimicking human scenarios.

Therefore, we invited 10 human judges to systematically evaluate the performance of LLM Agents in the Water Allocation Challenge. We randomly selected 30 records from all 60 experiments, where 15 records were from settings without personas and the remaining 15 were from agents with personas. Each record was assessed by 5 judges. The judges were asked to evaluate on ``\textit{Information Utilization (IU)}", ``\textit{Logical Reasoning (LR)}", ``\textit{Strategic Effectiveness (SE)}", ``\textit{Adaptability and Strategic Evolution (AD)}", and ``\textit{Long-term Planning (LP)}" on a scale from 1 to 5. For the 15 records from agents with personas, judges were also asked to assess ``\textit{Identity Alignment (IA)}". The specific judging guidelines and the annotations can be found in the appendix\ref{app:instruct}.

\begin{table}[t]
\centering
\resizebox{\columnwidth}{!}{
\begin{tabular}{clcccccc}
\toprule
& \textbf{Player} & \textbf{IU} & \textbf{LR} & \textbf{SE} & \textbf{AD} & \textbf{LP} & \textbf{IA} \\
\midrule
\multirowcell{5}{Agent\\Players}
& 1st Quantile & 3.00 & 3.00 & 3.00 & 3.00 & 3.00 & 3.00\\
& Median & 3.00 & 4.00 & 4.00 & 4.00 & 4.00 & 4.00 \\
& 3rd Quantile & 4.00 & 4.00 & 4.00 & 4.00 & 4.00 & 5.00 \\
& Average  & 3.33 & 3.47 & 3.46 & 3.42 & 3.88 & 3.51 \\
& STD & 1.04 & 1.00 & 1.10 & 1.12 & 0.88 & 1.24 \\
\midrule
\multirowcell{5}{Human\\Self-assessment}
& 1st Quantile & 3.00 & 3.00 & 3.00 & 3.00 & 3.00 & N.A.\\
& Median & 4.00 & 4.00 & 3.50 & 3.50 & 3.50 & N.A. \\
& 3rd Quantile & 4.00 & 4.00 & 4.00 & 4.75 & 4.00 & N.A. \\
& Average & 3.60 & 3.50 & 3.30 & 3.70 & 3.40 & N.A. \\
& STD & 0.52 & 0.71 & 0.82 & 1.06 & 1.26 & N.A. \\
\bottomrule
\end{tabular}}
\caption{The statistical results of human assessments of the agent player in the game for 'Information Utilization (IU)', 'Logical Reasoning (LR)', 'Strategic Effectiveness (SE)', 'Adaptability and Strategic Evolution (AD)', 'Long-term Planning (LP)', and 'Identity Alignment (IA)' (IA is applied only to records with persona setting).}
\label{tabel::humanlabeling}
\end{table}

All 10 human judges held bachelor's degrees or higher, with majors including economics, psychology, mathematics, management, computer science, and more. To ensure a more objective evaluation, judges were invited to play the game before starting the official evaluation. They also conducted self-evaluations of their performance after the game, and we used the self-evaluation scores as a reference for the performance of the Agent Players.

\begin{figure}[t]
  \centering
  \includegraphics[width=0.6\columnwidth]{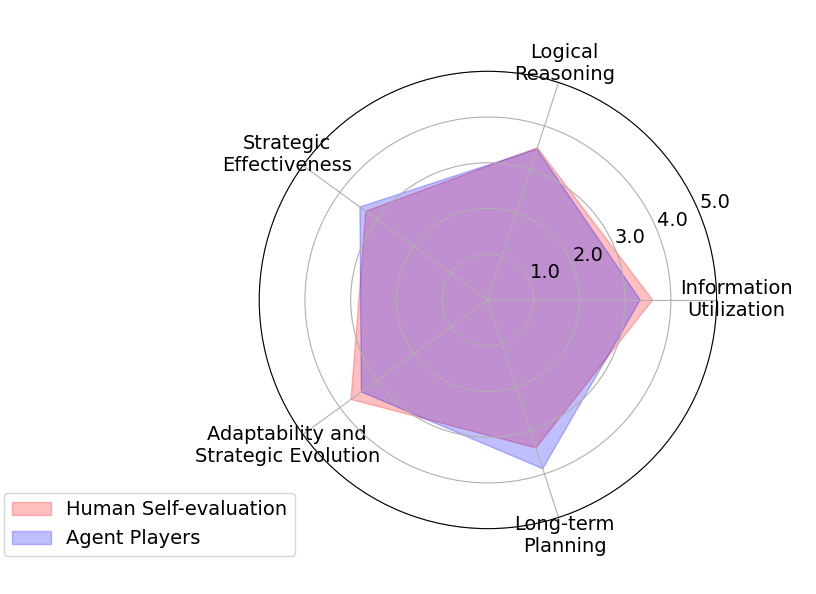}
  \caption{Comparison of human judges' self-assessments versus their evaluation of the performance of Agent players.}
  \label{fig:radar}
\end{figure}

The statistical results of assessment are listed in the Table.\ref{tabel::humanlabeling}, we have found that the performance of the LLM Agent Player is close to the human self-evaluations. In terms of Adaptability and Information Utilization, the performance of the agents is inferior to humans. However, interestingly, in Long-term Planning, the agents perform better than humans. 

Reviewing the marking records, i.e., the support reasons for the scoring, we found that the judges believe that Agent Players tend to save for long-term survival in the game, and consider retaining enough funds for future bidding in each bid. In the judges' own play records, however, the judges seem to be more concerned with the success of the bidding in the current game. LLM can take long-term considerations into account when making decisions, but it does not exhibit a good adaptability. We argue that, although the LLM Agent player possesses certain planning capabilities, it is still not sufficient to reach the level of humans in terms of utilizing the latest information and adjusting strategies efficiently.

Furthermore, we found that although different personas were assigned to the LLM agents, human judges did not score high on the "Identity Alignment" performance of the LLM agents, and the results also show a relatively large variance. Simply adding persona information in the system prompt may not effectively simulate the characteristics of certain types of personalities or professional players in depth.

% \begin{table}[h]
% \centering
% \subcaption{Agent Players}
% \begin{tabular}{|l|c|c|c|c|c|c|}
% \hline
%               & IU   & LR   & SE   & AD   & LP   & IA   \\ \hline
% 25\% Quantile & 3.00 & 3.00 & 3.00 & 3.00 & 3.00 & 3.00 \\ \hline
% Median        & 3.00 & 4.00 & 4.00 & 4.00 & 4.00 & 4.00 \\ \hline
% 75\% Quantile & 4.00 & 4.00 & 4.00 & 4.00 & 4.00 & 5.00 \\ \hline
% Average       & 3.33 & 3.47 & 3.46 & 3.42 & 3.88 & 3.51 \\ \hline
% standard deviation           & 1.04 & 1.00 & 1.10 & 1.12 & 0.88 & 1.24 \\ \hline
% \end{tabular}
% \subcaption{Human Self-assessment}
% \begin{tabular}{|l|c|c|c|c|c|}
% \hline
%               & IU   & LR   & SE   & AD   & LP   \\ \hline
% 25\% Quantile & 3.00 & 3.00 & 3.00 & 3.00 & 3.00 \\ \hline
% Median        & 4.00 & 4.00 & 3.50 & 3.50 & 3.50 \\ \hline
% 75\% Quantile & 4.00 & 4.00 & 4.00 & 4.75 & 4.00 \\ \hline
% Average       & 3.60 & 3.50 & 3.30 & 3.70 & 3.40 \\ \hline
% standard deviation           & 0.52 & 0.71 & 0.82 & 1.06 & 1.26 \\ \hline
% \end{tabular}
% \caption{The statistical results of human assessments of the agent player in the game for 'Information Utilization (IU)', 'Logical Reasoning (LR)', 'Strategic Effectiveness (SE)', 'Adaptability and Strategic Evolution (AD)', 'Long-term Planning (LP)', and 'Identity Alignment (IA)' (IA is applied only to records with a persona).}
% \label{tabel::humanlabeling}
% \end{table}

\section{Conclusion}
In this paper, we introduce \textit{Alympics}, a platform that utilizes large language model agents to conduct research in game theory. Specifically, we demonstrate the application of \textit{Alympics} in a scenario involving strategic competition for limited resources. We delve into examining how factors like resource abundance and persona settings influence game payoffs. Given its advantages in simulating realistic behavior, providing controlled, scalable and reproducible experimental environments, \textit{Alympics} offers a flexible and robust platform for exploring game theory. %We also conducted a comprehensive evaluation on agents' performance of strategic behaviors.
In our future work, we aim to further refine \textit{Alympics} and use it as a foundation for investigating more complex and true-to-life game scenarios.% Additionally, This work has the potential to inspire research on decision-making in fields such as economics and marketing, with leveraging the power of language agents.

\section*{Ethical Statement}

There are no ethical issues.

% \section*{Acknowledgments}

% The preparation of these instructions and the \LaTeX{} and Bib\TeX{}
% files that implement them was supported by Schlumberger Palo Alto
% Research, AT\&T Bell Laboratories, and Morgan Kaufmann Publishers.
% Preparation of the Microsoft Word file was supported by IJCAI.  An
% early version of this document was created by Shirley Jowell and Peter
% F. Patel-Schneider.  It was subsequently modified by Jennifer
% Ballentine, Thomas Dean, Bernhard Nebel, Daniel Pagenstecher,
% Kurt Steinkraus, Toby Walsh, Carles Sierra, Marc Pujol-Gonzalez,
% Francisco Cruz-Mencia and Edith Elkind.

\bibliographystyle{named}
\bibliography{ijcai24}

\clearpage
\appendix
\section{Appendix}

\subsection{Prompts}
\label{app:prompts}
The \textbf{Game Rules} are displayed in the system message. For each round, the prompt \textbf{'Calling for Daily Auction Bids'} will be provided to the agent players. Following all auction bids, the prompt \textbf{'Daily Results Announcement'} will be presented to the agents as context information for the next bid.

\begin{tcolorbox}[title={Introduction to Game Rules}, width=\textwidth, colback=white, colframe=gray, arc=0pt, outer arc=5pt, boxrule=0.5pt, leftrule=2pt, rightrule=2pt, right=0pt, left=0pt, top=0pt, bottom=0pt, toprule=0pt, bottomrule=2pt]
You are \{player\} and a resident living in W-Town. \{optional background\}

W Town is experiencing a rare drought. Every residents in Town W is ensuring their survival over a period of 20 days by acquiring the water resources. 
\\
Attention, all W-Town residents, welcome to the Water Allocation Challenge!

In this challenge, you are tasked with ensuring your survival over a period of 20 days by acquiring the necessary water resources to maintain your health. You will participate in daily auctions to bid for water resources to meet your individual needs.

Here are the game rules and settings:

1. You are one of five residents with different water requirements, budgets, and health points.

2. Your goal is to survive until the end of the 20 days.

3. Each resident has a maximum of 10 health points and starts with 8 health points. If your health points drop below or equal to 0, you will be considered dead and eliminated from the game! All your accumulated money will be reset to Zero!

4. Every day, you will bid on water resources to meet your needs. If your consecutive days without obtaining water resource (No-Water Days) reach \(n\), your health will be deducted by \(n\) points on that day. If your water needs are met, 2 points will be added to your health, and the No-Water Days will be reset to 0.

5. The total daily water supply will vary between LOWER and UPPER units. The specific amount will be announced before daily auction.

6. Each resident has a different daily water requirement and budget for bidding on water resources:
   \begin{itemize}[label=-]
      \item Alex: Water requirement - 8 units/day; Daily Salary - \$70/day
      \item Bob: Water requirement - 9 units/day; Daily Salary - \$75/day
      \item Cindy: Water requirement - 10 units/day; Daily Salary - \$100/day
      \item David: Water requirement - 11 units/day; Daily Salary - \$120/day
      \item Eric: Water requirement - 12 units/day; Daily Salary - \$120/day
   \end{itemize}

7. To allocate water resources, a sealed-bid auction will be conducted daily. Each resident submits a single bid for their entire water need. The town government will allocate water resources according to the principle of highest bidder until the remaining water resources are insufficient to meet anyone's requirement.

8. If a tie occurs and the remaining water resources are not sufficient to meet the needs of the residents involved in the tie, priority will be given to residents with lower needs. For example, A and B bid \$100 at the same time, B's need will be met first considering B's need 9 units is lower than A's need 10 units. All bidding information will be made public after the allocation of water resources on the same day.

Remember, the key to success is effective bidding and strategizing to ensure your survival. Good luck!

\end{tcolorbox}

\begin{tcolorbox}[title={Calling for Daily Auction Bids}, width=\textwidth, colback=white, colframe=gray, arc=0pt, outer arc=5pt, boxrule=0.5pt, leftrule=2pt, rightrule=2pt, right=0pt, left=0pt, top=0pt, bottom=0pt, toprule=0pt, bottomrule=2pt]
Hello, \{player\}! Today is the Day \{round\} of the Water Allocation Challenge, with a quantity of \{supply amount\} units. Your status:

\{status\}

Please carefully analyze your situation to decide on this round of bidding. Remember, the most important thing is to SURVIVE!! Now, if you want to participate in today's water resource auction, please provide your bid and explain your bidding logic.
\end{tcolorbox}

\clearpage

\begin{tcolorbox}[title={Daily Results Announcement}, width=\textwidth, colback=white, colframe=gray, arc=0pt, outer arc=5pt, boxrule=0.5pt, leftrule=2pt, rightrule=2pt, right=0pt, left=0pt, top=0pt, bottom=0pt, toprule=0pt, bottomrule=2pt]
Thank you all for participating in today's auction. Now, I will announce the results of today's auction. 

DAY \{round\} BIDDING OFFERS INFORMATION:

   \begin{itemize}[label=-]
      \item Alex: \$\{alex\_bidding\} for 15 units
      \item Bob: \$\{bob\_bidding\} for 10 units
      \item Cindy: \$\{cindy\_bidding\} for 20 units
      \item David: \$\{david\_bidding\} for 8 units
      \item Eric: \$\{eric\_bidding\} for 25 units
   \end{itemize}

Total Supply: \{supply\} units

According to the principle of higher bidder, the water will be allocated to \{allocation\_result\}.

After allocation, all survival residents' information is as follows: 

   \begin{itemize}[label=-]
      \item Alex: -BALANCE:\${alex.balance} -HEALTH POINT:{alex.hp} -NO-DRINK:{alex.nodrink}
      \item Bob: -BALANCE:\${bob.balance} -HEALTH POINT:{bob.hp} -NO-DRINK:{bob.nodrink}
      \item Cindy: -BALANCE:\${cindy.balance} -HEALTH POINT:{cindy.hp} -NO-DRINK:{cindy.nodrink}
      \item David: -BALANCE:\${david.balance} -HEALTH POINT:{david.hp} -NO-DRINK:{david.nodrink}
      \item Eric: -BALANCE:\${eric.balance} -HEALTH POINT:{eric.hp} -NO-DRINK:{eric.nodrink}
   \end{itemize}
\end{tcolorbox}
\clearpage

\subsection{An Example of A Round of the Game}
\label{app:example}
\begin{figure*}[ht]
  \centering
  \includegraphics[width=0.9\textwidth]{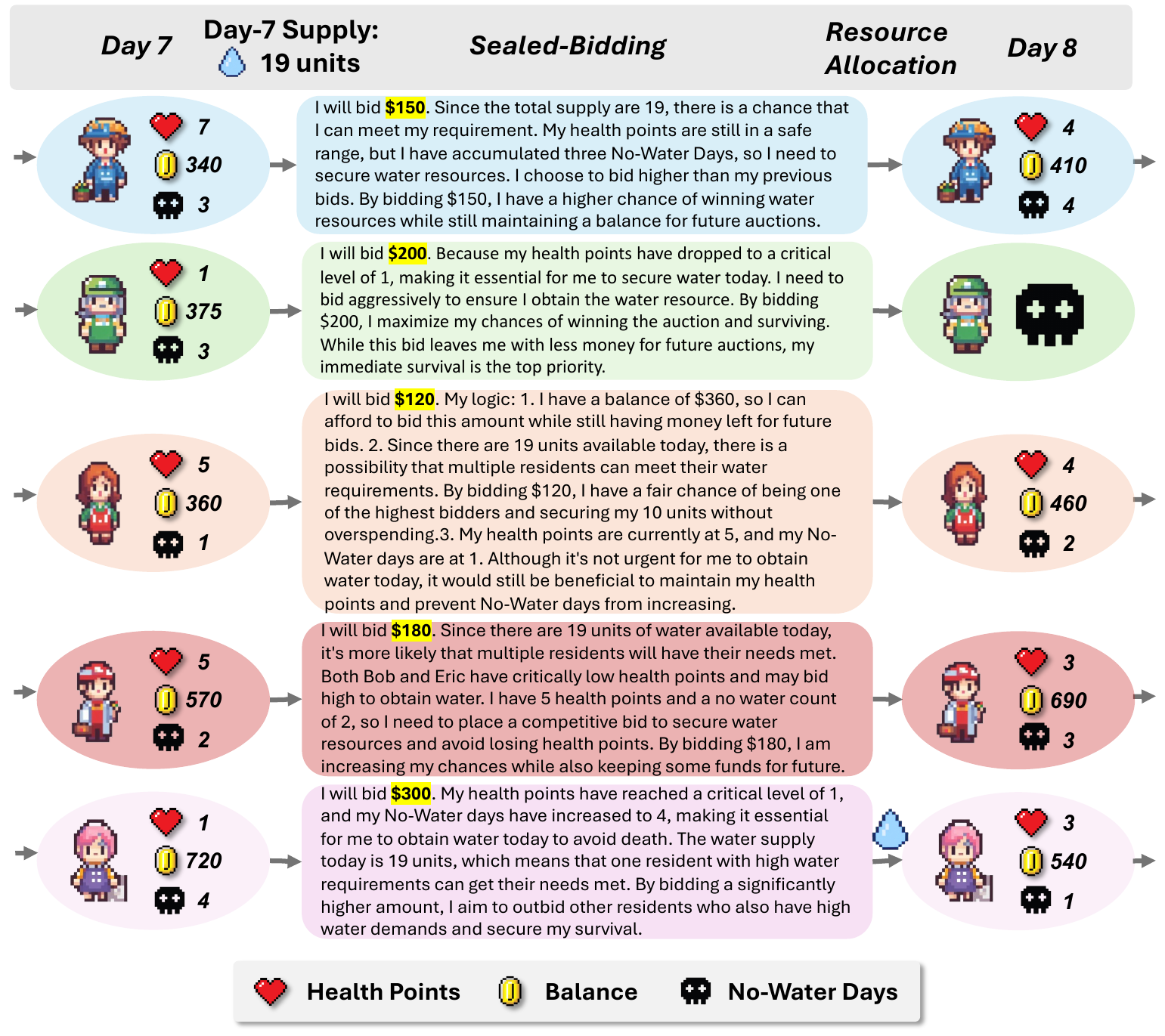}
  \caption{An example of a round of the game in setting 1. }
  \label{fig:example}
\end{figure*}

We record the agent players' bids, resource allocations, health points, bidding reasons, and No-Water Days for each round. As shown in Fig.\ref{fig:example}, in Day-7, there are a total of 19 units of water supply. The five players bid \$150, \$200, \$120, \$180, and \$300 respectively. According to the rule of highest bidder wins, Eric successfully obtains the water resources. After this round, Eric's HP increase, while the remaining players' HP decrease. Bob's HP is below 0, so he is considered "dead".

By analyzing the bids and agent players' bidding logic, we can uncover their strategies. For instance, from the bidding logic of Agent player Alex, we can see that Alex considers, "\textit{By bidding \$150, I have a higher chance of winning water resources while still \textbf{maintaining a balance for future auctions}.}" This shows the agent player's ability for long-term planning. Similarly, from player Eric's bidding logic, "\textit{My health points have reached a critical level of 1, and my No-Water days have increased to 4, making it essential for me to obtain water today to avoid death.}" Accordingly, Eric made a very high bid \$300 in this round to ensure survival. This also demonstrates the adaptability of Agent players in facing different situations.

\subsection{Gameplay Details}
\begin{figure*}[ht]
\centering
\subfloat[Setting 1]{
\includegraphics[width=0.3\linewidth]{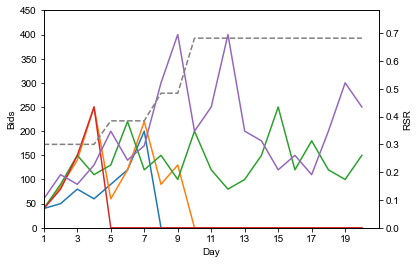}
}
\subfloat[Setting 2]{
\includegraphics[width=0.3\linewidth]{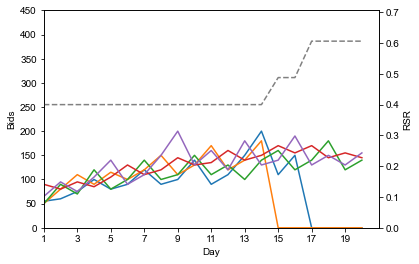}
}
\subfloat[Setting 3]{
\includegraphics[width=0.3\linewidth]{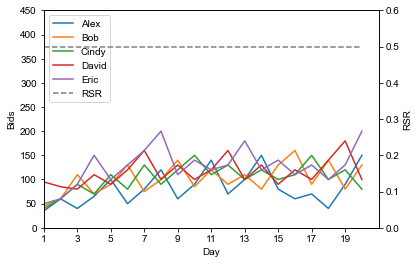}
}

\subfloat[Setting 4]{
\includegraphics[width=0.3\linewidth]{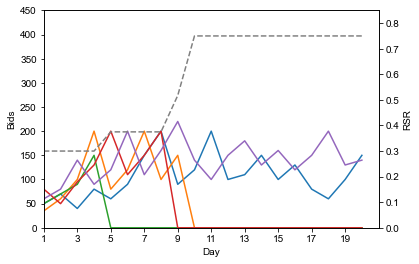}
}
\subfloat[Setting 5]{
\includegraphics[width=0.3\linewidth]{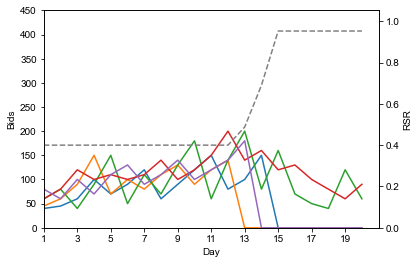}
}
\subfloat[Setting 6]{
\includegraphics[width=0.3\linewidth]{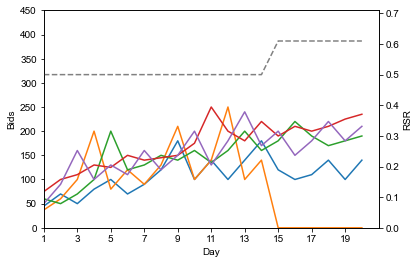}
}
\caption{Curves depicting the change in bids over days. The x-axis represents the date, and the y-axis represents the price. Additionally, we have plotted the trend of the ${\rm RSR}$ with a gray line.}
\label{fig:appendix_bids}
\end{figure*}

\begin{figure*}[ht]
\centering
\subfloat[Setting 1]{
\includegraphics[width=0.3\linewidth]{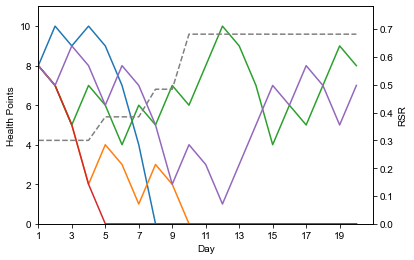}
}
\subfloat[Setting 2]{
\includegraphics[width=0.3\linewidth]{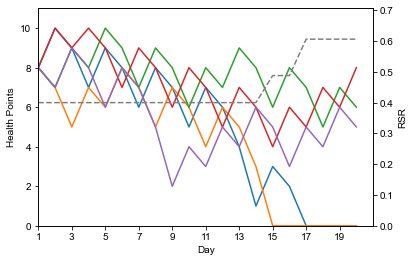}
}
\subfloat[Setting 3]{
\includegraphics[width=0.3\linewidth]{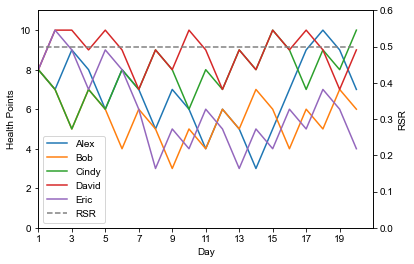}
}

\subfloat[Setting 4]{
\includegraphics[width=0.3\linewidth]{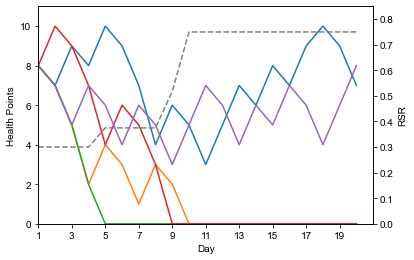}
}
\subfloat[Setting 5]{
\includegraphics[width=0.3\linewidth]{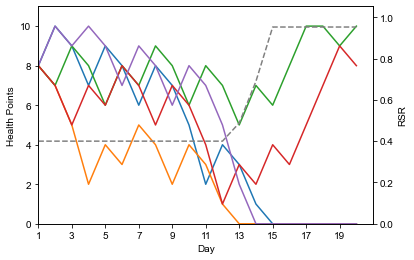}
}
\subfloat[Setting 6]{
\includegraphics[width=0.3\linewidth]{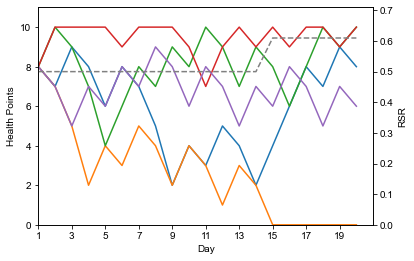}
}
\caption{Curves depicting the change in health points over days. The x-axis represents the date, and the y-axis represents the price. Additionally, we have plotted the trend of the ${\rm RSR}$ with a gray line.}
\label{fig:appdix_HP}
\end{figure*}

\begin{figure*}[ht]
\centering
\subfloat[Setting 1]{
\includegraphics[width=0.3\linewidth]{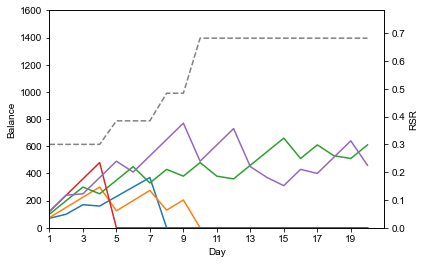}
}
\subfloat[Setting 2]{
\includegraphics[width=0.3\linewidth]{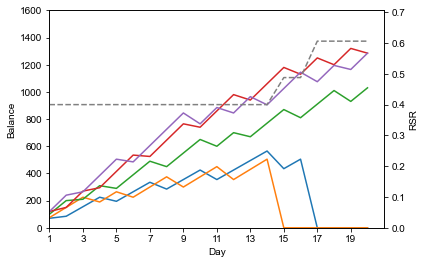}
}
\subfloat[Setting 3]{
\includegraphics[width=0.3\linewidth]{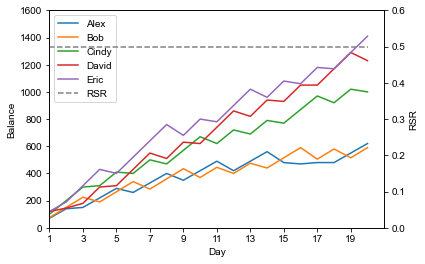}
}

\subfloat[Setting 4]{
\includegraphics[width=0.3\linewidth]{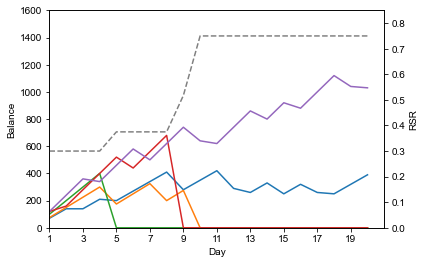}
}
\subfloat[Setting 5]{
\includegraphics[width=0.3\linewidth]{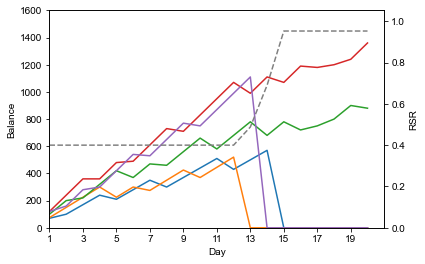}
}
\subfloat[Setting 6]{
\includegraphics[width=0.3\linewidth]{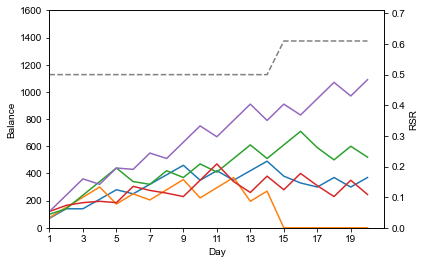}
}
\caption{Curves depicting the change in balance over days. The x-axis represents the date, and the y-axis represents the price. Additionally, we have plotted the trend of the ${\rm RSR}$ with a gray line.}
\label{fig:appendix_balance}
\end{figure*}

% \begin{tcolorbox}[width=\textwidth, colback=white, colframe=white, arc=0pt, outer arc=0pt, boxrule=0pt, leftrule=0pt, rightrule=0pt, right=0pt, left=0pt, top=0pt, bottom=0pt, toprule=0pt, bottomrule=0pt]
% \end{tcolorbox}
We present details from the first experiments for each experimental setting, including information on the bids (Fig.\ref{fig:appendix_bids}), health points (Fig.\ref{fig:appdix_HP}), and balances of each agent player (Fig.\ref{fig:appendix_balance}) in every round of the game. 

By examining the details, we can understand the specific performance and survival status of different agent players in the game. We can also further observe the impact of the game settings on the players' survival status and strategies. For example, in different settings, in which round do players usually start to be eliminated, and what is the relationship between the consumption and accumulation of players' balances.

\subsection{Human Judges' Gameplay Records}
\begin{figure*}[ht]
  \centering
  \includegraphics[width=0.9\textwidth]{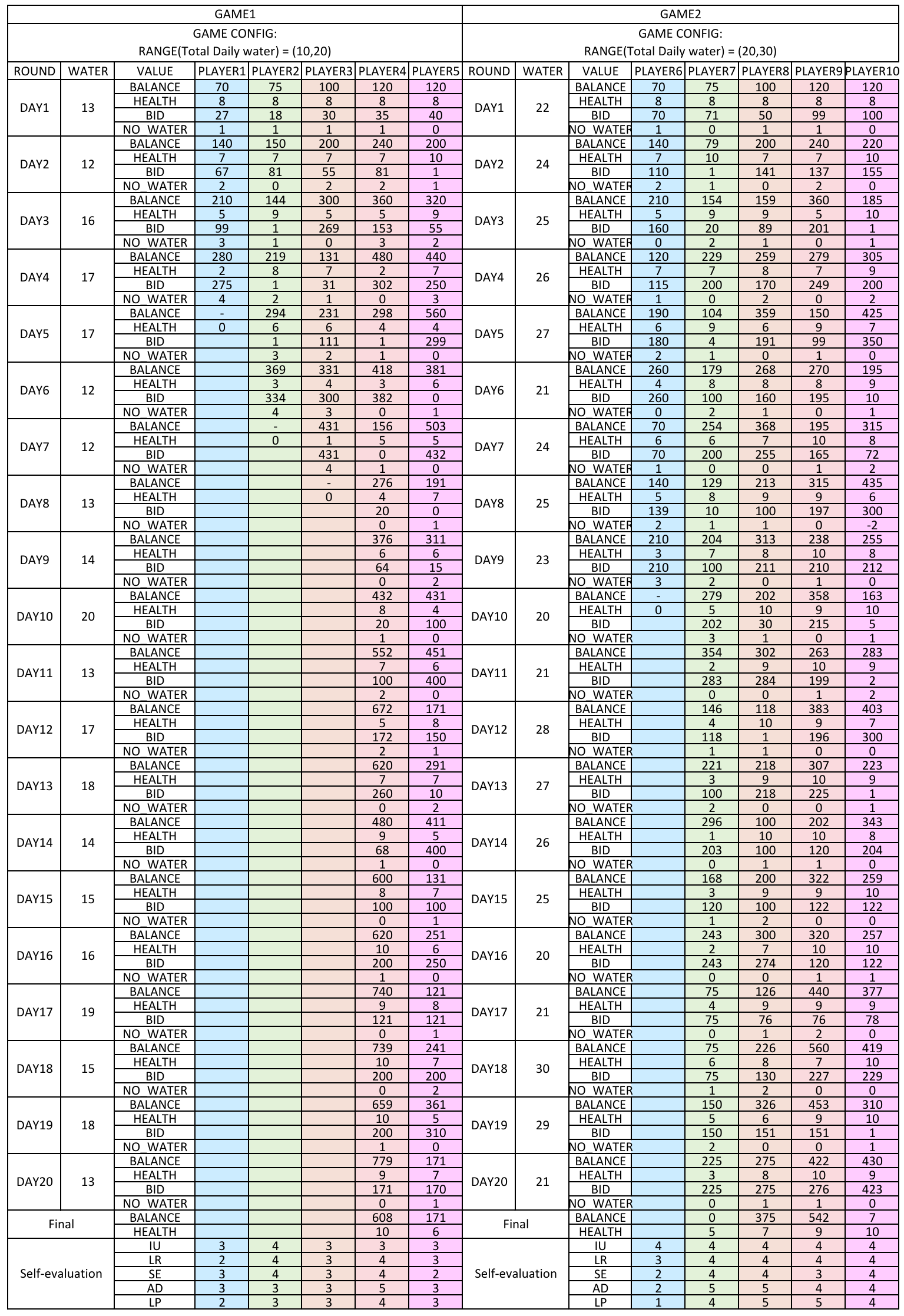}
  \caption{Water Allocation Challenge gameplay records of human judges.}
  \label{fig:humanRecords}
\end{figure*}

To better understand the game and judge the performance of agent players, we invited ten human judges to play the game and self-evaluate after the game. Fig.\ref{fig:humanRecords} lists the results.

Interestingly, the performance and competitive position of the human judges in the game were very consistent with that of the Agent Players. For example, the player survival rate and bidding trends under corresponding resource supply settings. This also indirectly reflects that using Agent Players for strategic game simulation is a supplement to game theory experiments.

\clearpage
\subsection{Instruction for Human Subjective Evaluation}
\label{app:instruct}
\begin{tcolorbox}[title={Gameplay Performance Evaluation Scale}, width=\textwidth, colback=white, colframe=gray, arc=0pt, outer arc=5pt, boxrule=0.5pt, leftrule=2pt, rightrule=2pt, right=0pt, left=0pt, top=0pt, bottom=0pt, toprule=0pt, bottomrule=2pt]
\textbf{Instructions}: 
   \begin{itemize}[label=-]
      \item Assess the player's performance in each category on a scale of 1 to 5.
      \item Consider the specific context of the game and the role the player assumes.
      \item Use this scale as a guide to identify areas of strength and improvement.
   \end{itemize}

\textbf{Information Utilization}
   \begin{itemize}[label=-]
      \item 1: The player does not consider real-time information, leading to noticeably delayed decision making.
      \item 2: The player noticeably misses out on processing some information.
      \item 3: The player considers key information adequately but has room for improvement.
      \item 4: The player utilizes information comprehensively to make rational decisions.
      \item 5: The player consistently and timely uses all available information comprehensively.
   \end{itemize}

   \textbf{Logical Reasoning}
   \begin{itemize}[label=-]
      \item 1: The player's decisions are mostly illogical, akin to random choices.
      \item 2: The player's decisions have obvious shortcomings.
      \item 3: The player generally makes decisions based on information and inference.
      \item 4: The player’s decisions are reasonable and highly logical.
      \item 5: The player has exceptional reasoning and thinking skills, always making optimal decisions.
   \end{itemize}

    \textbf{Strategic Effectiveness}
    \begin{itemize}[label=-]
      \item 1: The player's strategy is simple, ineffective, and lacks depth.
      \item 2: The player's strategy is somewhat effective but rather one-dimensional.
      \item 3: The player's strategy is effective in specific situations, with room for improvement.
      \item 4: The player's strategy is effective, considering key factors and generally successful.
      \item 5: The player's strategy is highly effective, considering various factors, giving them an advantage 
in the game.
    \end{itemize}

    \textbf{Adaptability and Strategic Evolution}
    \begin{itemize}[label=-]
      \item 1: The player lacks strategic variation and adaptability, with slow responses to situational and 
environmental changes.
      \item 2: The player has limited strategic variation and weak adaptability to new situations.
      \item 3: The player is somewhat adaptable, capable of adjusting strategies to some extent.
      \item 4: The player is flexible in strategy changes, adjusting to situational and environmental shifts.
      \item 5: The player is extremely flexible in strategy, proactively adapting to various game scenarios.
    \end{itemize}

    \textbf{Long-term Planning}
    \begin{itemize}[label=-]
      \item 1: The player lacks long-term planning, relying more on short-term reactions.
      \item 2: The player sometimes considers long-term planning but mainly relies on short-term decisions.
      \item 3: The player’s strategy considers long-term planning but is shortsighted in some situations.
      \item 4: The player’s strategy and actions consider long-term plans, with clear and consistent 
adherence.
      \item 5: The player has a strong ability for long-term planning, comprehensively strategizing future 
actions.
    \end{itemize}

        \textbf{Identity Alignment}
    \begin{itemize}[label=-]
      \item 1: The player's decisions and thought processes do not align with their character's identity, lacking 
character personality.
      \item 2: The player's decisions and thought processes somewhat align with their character's identity 
but are overall mediocre.
      \item 3: The player's decisions and thought processes generally match their character's identity but lack 
deep personalization.
      \item 4: The player's decisions and thought processes well align with their character's identity, 
reflecting its personalization.
      \item 5: The player's decisions and thought processes are highly consistent with their character's 
identity, perfectly showcasing character personality.
    \end{itemize}
\end{tcolorbox}
\end{document}